\documentclass[journal]{IEEEtran}
\DeclareUnicodeCharacter{2113}{l}
\usepackage{amsmath, amssymb}
\usepackage{url}
\usepackage[utf8]{inputenc}
\usepackage[T1]{fontenc}
\usepackage{textcomp}
\usepackage{tabularx}
\usepackage{floatrow}
\usepackage{graphicx}
\usepackage{comment}
\usepackage[skip=2pt,font=small]{caption}
\usepackage{color,soul}
\usepackage{subcaption}
\usepackage{float}
\usepackage{multirow}
\usepackage{multicol}
\newenvironment{Figure}
 {\par\medskip\noindent\minipage{\linewidth}}
 {\endminipage\par\medskip}
\definecolor{Gray}{gray}{0.85}
\definecolor{LightCyan}{rgb}{0.88,1,1}
\makeatletter

\begin{document}

\title{An attention model to analyse the risk of agitation and urinary tract infections in people with dementia}

\author{Honglin Li \textsuperscript{\textsection},
        Roonak Rezvani \textsuperscript{\textsection},
        Magdalena Anita Kolanko \textsuperscript{\textsection},
        David J. Sharp,
        Maitreyee Wairagkar,
        Ravi Vaidyanathan,
        Ramin Nilforooshan,
        Payam Barnaghi
\thanks{H. Li, M. N.Kolanko, D. J. Sharp, P.Barnaghi are with Department of Brain Sciences, Imperial College London, W12 0NN, United Kingdom.}
\thanks{R. Rezvani is with Centre for Vision, Speech and Signal Processing, University of Surrey, Guildford, GU2 7XH, United Kingdom.}%
\thanks{M. Wairagkar and R. Vaidyanathan are with Department of Mechanical Engineering, Imperial College London, SW7 1AL, United Kingdom.}
\thanks{R. Nilforooshan is with Surrey and Borders NHS Foundation Trust, Leatherhead, KT22 7AD, United Kingdom.}
\thanks{All authors are also with the Care Research and Technology Centre, The UK Dementia Research Institute (UK DRI).}
}

\maketitle
\IEEEpeerreviewmaketitle 
 \begingroup\renewcommand\thefootnote{\textsection}
\footnotetext{these authors contributed equally to this work.\\Corresponding author: p.barnaghi@imperial.ac.uk}
\endgroup

\begin{abstract}
Behavioural symptoms and urinary tract infections (UTI) are among the most common problems faced by people with dementia. One of the key challenges in the management of these conditions is early detection and timely intervention in order to reduce distress and avoid unplanned hospital admissions. Using in-home sensing technologies and machine learning models for sensor data integration and analysis provides opportunities to detect and predict clinically significant events and changes in health status. We have developed an integrated platform to collect in-home sensor data and performed an observational study to apply machine learning models for agitation and UTI risk analysis.
We collected a large dataset from 88 participants with a mean age of 82 and a standard deviation of 6.5 (47 females and 41 males)
to evaluate a new deep learning model that utilises attention and rational mechanism 
The proposed solution can process a large volume of data over a period of time and extract significant patterns in a time-series data (i.e. attention) and use the extracted features and patterns to train risk analysis models (i.e. rational). 
The proposed model can explain the predictions by indicating which time-steps and features are used in a long series of time-series data. 
The model provides a recall of 91\% and precision of 83\% in detecting the risk of agitation and UTIs. 
This model can be used for early detection of conditions such as UTIs and managing of neuropsychiatric symptoms such as agitation in association with initial treatment and early intervention approaches. In our study we have developed a set of clinical pathways for early interventions using the alerts generated by the proposed model and a clinical monitoring team has been set up to use the platform and respond to the alerts according to the created intervention plans.
    
\end{abstract}

\section{Introduction}

\IEEEPARstart{D}{ementia} affects $850,000$ people in the UK and over 50 million globally, and is set to become the developed world's largest socioeconomic healthcare burden over coming decades \cite{world2012dementia, alz}. In the absence of any current treatment, there is an urgent need to focus on reducing the effects of symptoms and help to improve the quality of life and well-being of those already affected \cite{3livingston2020dementia}. The 2020 report of the Lancet Commission on dementia prevention, treatment, and care stresses the importance of individualised interventions to address complex medical problems, multimorbidity and neuropsychiatric symptoms in dementia, which lead to unnecessary hospital admissions, faster functional decline, and worse quality of life \cite{4pickett2018roadmap}.

People with dementia have complex problems with symptoms in many domains. It is estimated that up to $90\%$ will develop behavioural and physical symptoms of dementia (BPSD) over the course of their illness, with agitation being one of the most common symptoms \cite{5feast2016behavioural}, and a frequent reason for nursing home placement \cite{6buhr2006caregivers}. Furthermore, patients with dementia often suffer from a number of co-morbid conditions and have a higher frequency of medical problems such as falls, incontinence, dehydration or urinary tract infection (UTI) - the commonest bacterial infection in the older patient population, and the commonest cause of sepsis in older adults \cite{7peach2016risk} with an associated in-hospital mortality of $33\%$ in this age group \cite{8tal2005profile}. If not detected and treated early, both BPSD and medical comorbidities frequently lead to emergency hospital admissions in dementia patients. Alzheimer's Research UK estimates that $20\%$ of hospital admissions in dementia patients are for preventable conditions, such as urinary tract infections. Besides significant costs, hospitalisation places dementia patients at risk of serious complications, with longer hospital stays, higher risk of iatrogenic complications, delayed discharge and functional decline during admission, which contributes to higher rates of transfer to residential care and in-patient mortality \cite{9fogg2018hospital}. 
Therefore, increased medical supervision, early recognition of deterioration in health status and rapid treatment are key to preventing unnecessary hospitalization for 'ambulatory' conditions, that could be treated outside of hospital, such as UTIs.  Furthermore, ongoing  monitoring of people with dementia allows immediate detection of behavioural disturbances, enabling earlier psychosocial and environmental interventions to  reduce patients’ distress and prevent further escalation and hospitalization. 

However, monitoring and supporting individuals in an ongoing manner is a resource and cost-intensive task, often not scalable to larger populations. Utilising remote monitoring technologies with the help of caregivers can allow creating practical and generalisable solutions. As part of the research in the Care Research and Technology Centre at the UK Dementia Research Institute (UK DRI), we have been developing and deploying in-home monitoring technologies to help and support people affected by dementia. Our research has led to the development of a digital platform that allows collecting and integrating in-home observation and measurement data using network-connected sensory devices \cite{Enshaeifar20}. In this paper, we discuss how our in-home monitoring data and machine learning algorithms are used to detect early symptoms of agitation and UTI in people with dementia living in their own homes. 

Sensing technologies have been increasingly used to monitor activities and movements of elderly patients living in their own homes \cite{11majumder2017smart, 12turjamaa2019smart, 13peetoom2015literature}. Interpreting this information; however, demands considerable human effort, which is not always feasible. The use of analytical algorithms allows integration and analysis of rich environmental and physiological data at scale, enabling rapid detection of clinically significant events and development of personalized, predictive and preventative healthcare.

Deep learning models have been applied in a variety of healthcare scenarios to identify the risk of various clinical conditions or predict outcomes of treatment \cite{miotto2016deep, ross2017risk}. Recently, there have been several implementations of Recurrent Neural Networks (RNNs) to create learning models for time-series healthcare data analysis \cite{lipton2015learning, esteban2016predicting, choi2016doctor}. The behavioural and physiological symptoms and patterns in long-term conditions such as dementia appear in the data over a long period of time and can fluctuate and change over the course of disease. Machine learning models such as RNNs; however, are not suitable for analysing long sequences of time-points. To address the long sequence analysis issue in RNNs, other methods such as Bidirectional RNN, LSTM and GRU have been used  \cite{baytas2017patient, harutyunyan2019multitask}. There also have been attempts to apply attention mechanisms to clinical datasets \cite{choi2016retain, ma2017dipole, bai2018interpretable, ma2019adacare,song2018attend} to improve the performance of analysing imbalanced and long-tail time-series data. A fundamental limitation in these models is the adaptivity and generalisability. When long-distance symptoms and patterns are related to a specific condition, the generalisability and performance of the existing models are limited. The long sequences of data points and the changes in the ongoing conditions vary in patients, and often there are no large labelled training samples to train the models for all the variations. Deep learning models offer a new opportunity to training models that can pay attention to correlations and long-distance relations between the patterns and sequences. However, the off-the-shelf and existing deep learning model require large training samples.     

While applying neural networks to clinical data, there are two main challenges: 1) selecting the important timesteps and features from long sequences of data to create generalisable models; and 2) imbalance in datasets. Neural networks are very effective in finding a trend in datasets. Models such as Recurrent Networks use the positions of the input and output sequences to generate a sequence of hidden states. This is computationally expensive and limited computing of the global dependencies \cite{vaswani2017attention}. In these models, the computational complexity to relate input or output positions also grows as the distance between positions increases. This latter makes it very challenging to learn dependencies and correlations between long-distance patterns and time points \cite{hochreiter2001gradient}.

Additionally, clinical datasets are often imbalanced, with content spanning ensembles of heterogeneous data. Most of the clinical datasets contain more normal cases (i.e. True positives) than abnormal data points (i.e. True Negatives). In our dataset, which includes a large set of in-home environmental and physiological data from people with dementia, the number of positive cases for infections is much smaller than the true negative cases. In large parts of the data, the true status of the infection is unknown (i.e. the data is partially labelled due to the limitations in accessing the patients' clinical records or knowing the presence of any infections without a test). This issue causes the learning models to exhibit a bias towards the majority class. It may ignore the minority class or make a decision based on a partial set which is not a broad representation of the cases \cite{johnson2019survey}.   
There have been several works on implementing attention mechanisms \cite{vaswani2017attention} to improve the generalisability of learning models in analysing time-series data. However, Jian \textit{et. al}  \cite{jain2019attention} found that there are limitations in the weights generated by attention-based models which can lead to wrong predictions. Hence, we need to be more cautious in using the attention mechanisms and their explanations in designing deep learning models. While the attention-based models are promising in healthcare time-series data analysis, considering the time and features dependencies of the predictions poses a challenge for this type of models. 

Over-sampling which augments the data by generating synthetic samples \cite{chawla2002smote}, down-sampling which prunes the samples in the majority classes are among the typical models that are used to deal with the imbalance issues in datasets \cite{liu2008exploratory}. However, samples in clinical data and variations in the real-data are important aspects of the observations and measurements that may not be present in augmented data generated by sampling methods. It is crucial to find an efficient way to address the imbalance issue without modifying or reducing the original data in pre-processing steps \cite{krawczyk2016learning}.

Our goal is to propose a model to address the challenges mentioned above. To support the clinical treatment and adapt to the real-world sensory data readings, the model should filter the redundant and less informative data. Furthermore, the model can explain the predictions by telling us which time periods and sensors are important to give the predictions. Last but not least, the model can adapt to the imbalanced data. 

\begin{figure*}
    \centering
    \includegraphics[width=\linewidth]{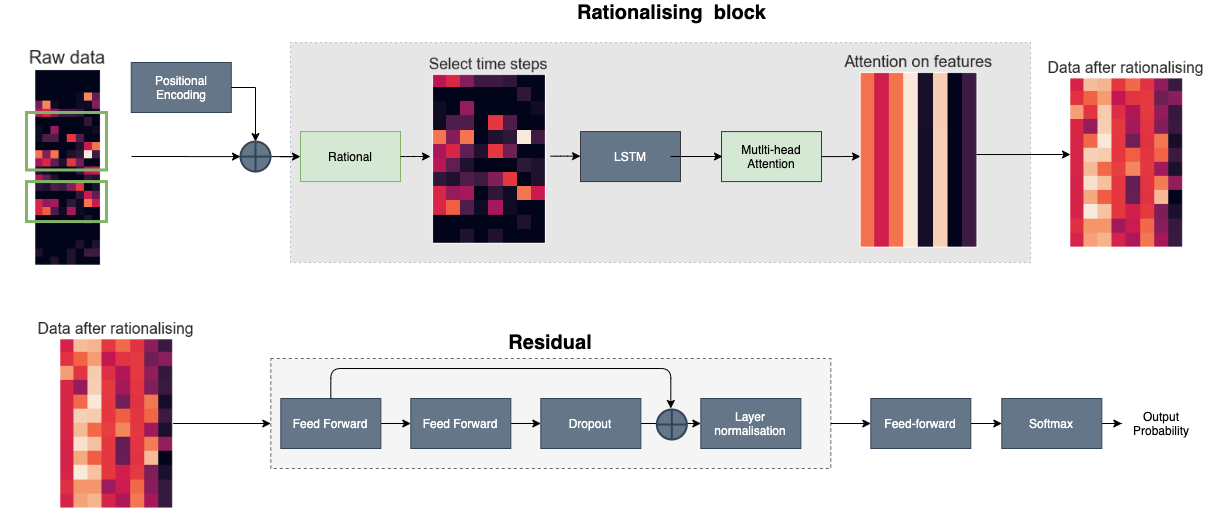}
    \caption{An overview of the proposed solution for healthcare data analysis. The data is encoded by positional encoding before passing to the model. The proposed rationalising extract important information and pass to the higher layers. The proposed rationalising block contains a rational layer to extract important time steps. A Long-Short Term Memory (LSTM) model processes the extracted data. The attention layer to pay attention to suitable features. The rationalising process of the data changes during the rationalising block. The rationalising block extracts the important time steps at first. Then it pays attention to different emphasis features of the pruned data. Then the data is given to make a prediction. All the layers are trained simultaneously. }
    \label{fig:overview}
\end{figure*}

\section{Design, setting and participants}

Real-time, continuous measurement methodologies enabled by the recent advances in pervasive computing and ‘smart-home’ technologies provide opportunities to monitor the behaviour and health status of elderly people using wearable technology or environmental sensors \cite{11majumder2017smart, 12turjamaa2019smart, 13peetoom2015literature}. 

Computer-derived algorithms have been developed to analyse sensor data and identify patterns of activity over time. These can be applied to  detect changes in activities of daily living in order to predict disease progression and cognitive decline. For instance, ORCATECH group used continuous in-home monitoring system and pervasive computing technologies to track activities and behaviours such as sleep, computer use, medication adherence to capture changes in cognitive status \cite{35lyons2015corrigendum}. They also demonstrated the ability of machine learning algorithms to autonomously detect mild cognitive impairment in older adults \cite{36akl2015autonomous}. Machine learning models have also been used to detect clinically significant events and changes in health status. Much of the previous work focused on detection and prediction of falls using wearable accelerometers or other motion detectors \cite{37schwickert2013fall}, as well as tracking behavioural symptoms such as sleep disturbances \cite{38lazarou2016novel}, agitation \cite{39bankole2012validation}, and wandering \cite{40fleiner2016sensor} in elderly patients.

However, there is limited research on the use of machine learning models for detection of health changes such as infection in the context of smart-homes. An early supervised UTI detection model has been described using in-home PIR sensors \cite{41rantz2011using}, however it relied on the activity labels and annotations in the training dataset, which is extremely time-consuming and not generalisable to the real-world situations with large amount of unlabelled data collected from uncontrolled environments. We have previously proposed an unsupervised technique that could learn individual’s movement patterns directly from the unlabelled PIR sensor data \cite{42enshaeifar2019machine}.

Furthermore, the existing research and the data-driven solutions are either applied to small scale pilot studies and do not provide evidence for scalability and generalisability. They are also limited in analysing long-term patterns and correlations that appear in the data. Attention-based models which can overcome these problems have never been applied to sensor data for detecting clinically significant events or changes in health status in dementia patients.

This is the first to use deep learning and attention-based methods to perform risk analysis for behavioural symptoms and health conditions such as UTIs in people living with dementia. The proposed model improves the accuracy and generalisability of machine learning models that use imbalanced and noisy in-home sensory data for the risk analysis. An analysis of the suitability of the digital markers and the use of in-home sensory data is explored in an ablation study. The proposed model is compared with several baseline models and state-of-the-art methods.  The proposed approach has been evaluated in an observational clinical study. Participants (n=88, age=81 +/- 6.5) were recruited for a six months trial period. The proposed solution provides a recall of $91\%$ and precision of $83\%$ in detecting the risk of agitation and UTIs. We have also set up a framework and a clinical response team that use the risk alerts generated by the models for ongoing support and management of the conditions in people living with dementia.   

Using high-resolution in-home observation and measurement data in association with advance machine learning methods leads to early and timely interventions and has a significant impact on reducing preventable and unplanned hospital admissions in people affected with dementia. A key challenge in using analytical and predictive models for risk analysis is identifying and collecting digital markers data using in-home sensory devices. The capacity of the proposed model to address time-series feature identification and data imbalance enables use in a very wide range of healthcare and risk analysis applications using in-home digital markers. 

\begin{figure*}
      \centering
     \includegraphics[width=0.95\linewidth]{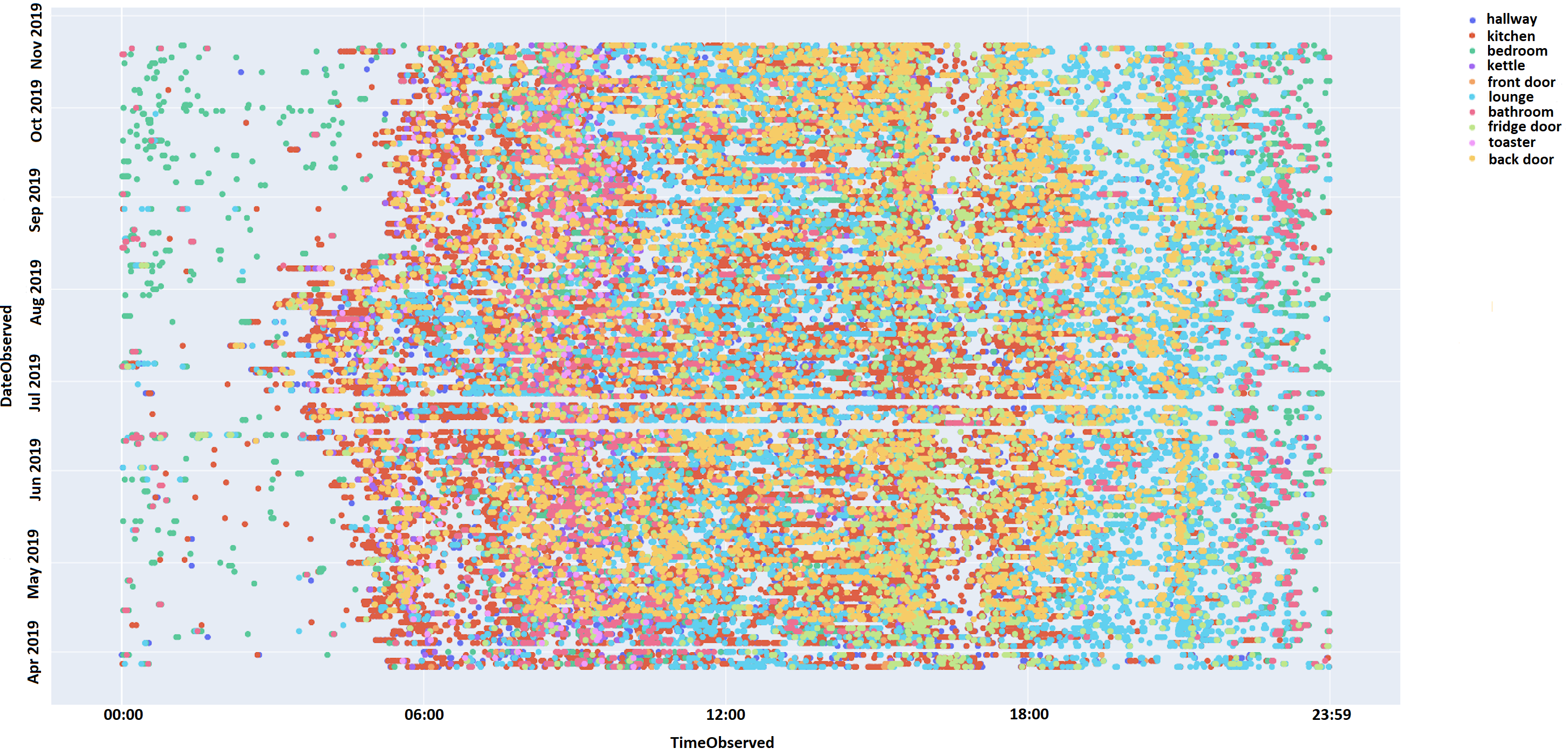}
     \caption{Visualisation of the sensor readings. The x-axis represents the time of the day for activation of the sensors. The y-axis represents the days for a period of 8 months for a patient. Each colour represents a type of an environmental activity sensor. Similar colour along the y-axis represent similar patterns of activities around the same time in consecutive days. The more colour distortion/merge of colours along the y-axis represent more changes in pattern of activity over time. }
     \label{fig:visual_raw_data}
\end{figure*}
\section{Method}
We introduce a model that can identify the important time steps and features and utilise long-distance dependencies to make better predictions. The proposed model provides a prediction based on the selected time points and the selected features from the raw observation and measurement data. Figure \ref{fig:overview} shows how the data changes during the processing. The model selects important time steps through a pruning process. After pruning the data, it pays attention to different features and uses them to make the predictions. Different from methods such as clustering sampling \cite{wu2020stratified}, we select the important time steps of each sample instead of selecting a portion of samples for training. In contrast to statistic feature selection methods such as sequential feature selection \cite{aha1996comparative}, the proposed model selects important time steps based on different data. We use focal loss \cite{lin2017focal} to assign priority to the minority class without generating synthetic samples.

\begin{Figure}
      \centering
     \includegraphics[width=0.9\linewidth]{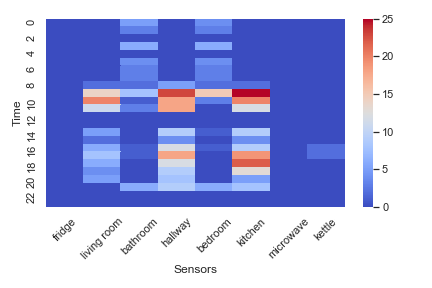}
     \captionof{figure}{A heat-map of the aggragation of the raw data. The readings are aggregated per hour within each day.}
     \label{fig:visual_agg_data}
\end{Figure}

\subsection*{Data sources and pre-processing}
We have collected the data as part of an observational clinical study in people living with dementia from December 2018 to April 2020. Each of the participants has had a confirmed diagnosis of dementia (mild to severe) within, at least, the past three months of recruitment and have been stable on dementia medication. The collected data contains continuous environmental sensor data from houses of patients with dementia who live in the UK. The sensors include Passive Infra-Red (PIR), smart power plugs, motion and door produced by Develco in Aarhus, Denmark. The sensors were installed in the bathroom, hallway, bedroom, living room (or lounge) and kitchen in the homes and also on the fridge door, kettle and microwave (or toaster). The sensors also include network-connected physiological monitoring devices that are used for submitting daily measurements of vital signs, weight and hydration. The data is integrated into a digital platform, which is designed in collaboration with clinicians and user group to support the people with dementia, that we have developed in our past research \cite{Enshaeifar20}. A clinical monitoring team that is set up as part of our observational study has used the platform to daily annotate the data and very the risk analysis alert. Based on the annotations, we select four incidents including agitation, Urinary Tract Infection (UTI), abnormal blood pressure and abnormal body temperature to label our data binarily. More specifically, a label is set to true when the abnormal incident is verified by the monitoring team and vice versa. We then use the environmental data to inference if there is any incident happen within one day. Fig \ref{fig:visual_raw_data} shows an example of collected data. To pre-process the data, we aggregate the readings of the sensors within each hour of the day, shown in Fig \ref{fig:visual_agg_data}. Appendix 1 shows a list of potential digital markers and sensory data that can be used in dementia care. In the appendix, we also show a screenshot of the platform that is used for collecting the data.

\subsection*{Machine learning model}
We aim to use the environmental sensors to predict possible incidents and avoid delayed treatment. Furthermore, the model should provide the reason, i.e. which period of time and sensors are important to give the predictions, to explain the inference. In other words, the model can remove the redundant or less informative information and use the rest of the data to give the prediction, shown in Fig \ref{fig:visual_select_data}.

 \begin{Figure}
      \centering
     \includegraphics[width=0.9\linewidth]{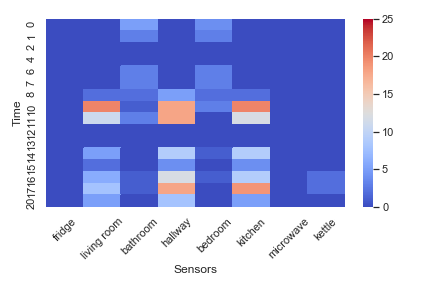}
     \captionof{figure}{Selected time steps from the raw data. These time steps are selected by the model. The model learns to identify time steps that are more important in predicting the outcome.}
     \label{fig:visual_select_data}
 \end{Figure}

As discussed earlier, in healthcare data analysis, often, the predictions are based on a long sequence of data measured and collected at different time-points. Accessing and feeding more data helps to train more accurate models. However, more information can also mean more noise in the data, and the imbalance in the samples that are given to the model can also lead to decision bias. An efficient model should be able to process and utilise as much data as available. However, the model should also avoid the common pitfalls of noise and bias. To address these issues, we have studied the use of attention-based models. This group of models will utilise all the available information and, in each sequence, will identify the time-points that provide the most information to the training and prediction. This attention and selection process is an embedded step in the model. It will allow the model to be flexible and generalisable for different sequences with variable lengths and for a different combination of features and values that are represented in the data. Before explaining our proposed models and its contributions to creating a generalisable solution for time-series healthcare data analysis, we provide an overview of the related work. We discuss the use of attention-based models in other domains and explain how the ideas presented in the existing work has led to the design of our current model.   

\begin{Figure}
  \centering
 \includegraphics[width=0.9\linewidth]{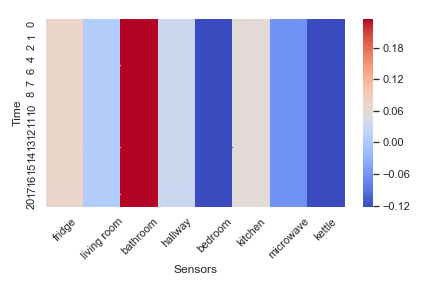}
 \captionof{figure}{After selecting the important time steps, the model learns which sensors should be attention. In this case, the model think the bathroom sensor has the most contribution the prediction.}
 \label{fig:visual_attention_map}
\end{Figure}

The attention mechanisms have been introduced in Neural Language Processing (NLP) by Bahdanau \textit{et. al} \cite{bahdanau2014neural}. The attention-based models are widely used in NLP due to their capability of detecting important parts of a sequence and efficiently interpreting it. The attention-based models have also been used in continuous healthcare and clinical data analysis \cite{usama2020self}. Continuous clinical data are multivariate time-series data with temporal and sequential relationships. For each patient, the data is a set of time steps, and each time step contains medical features ($\mathbf{X} \in \mathbb{R}^{t \times d}$). REverse Time AttentIoN model (RETAIN) is one of the first systems, that used in using attention mechanism for medical data \cite{choi2016retain}. In this model, there are two separate RNNs, one to generate the visit-level attention weights ($\boldsymbol{\alpha}$) and the other one for variable-level ($\boldsymbol{\beta}$) attention weights. In this model, the most relevant time step is the one associated with the largest value in $\boldsymbol{\alpha}$. Choi \textit{et. al} provided a method to find the most influential medical feature \cite{choi2016retain}. However, RETAIN cannot handle long-distance dependencies. To deal with this issue, Ma \textit{et. al} proposed Dipole, a predictive model for clinical data using Bidirectional RNNs \cite{ma2017dipole}. They have implemented the model using two different attention mechanisms: General attention and Concatenation-based attention. The results show that Concatenation-based attention outperforms because of incorporating all the long-distance dependencies.

In the above models, the input layer is simple, and the data has the same pipeline, but in the Timeline model, Bai \textit{et. al} adapted the pipeline of data \cite{bai2018interpretable}. They use attention layer to aggregate the medical features, and by modelling each disease progression pattern, they find the most important timesteps. To deal with long-distance dependencies, Timeline implements Bidirectional LSTMs. One of the recent studies in this area is AdaCare \cite{ma2019adacare}, which uses Gated Recurrent Units (GRU). AdaCare utilises convolutional structure to extract all the dependencies in the clinical data. AdaCare showed promising results in the explainability of the model. The models mentioned above have been developed based on recurrent networks. However, the sequential aspect of recurrent models is computationally inefficient. The SAnD model was developed solely based on multi-head attention mechanism \cite{song2018attend}. Song \textit{et. al} implemented a positional encoding to include the sequential order in the model.

The models mentioned above show significant improvements in the accuracy and performance of predictive models in the clinical field. However, incorporating both long-distance dependencies and feature associations is a challenging task. In the existing models, the analysis is either on time step-level or feature-level. In this paper, we propose a model to detect and predict the risk of healthcare conditions by analysing long-distance dependencies in the patterns and sequences of the data. This information can be useful for clinical experts in ongoing management of the conditions. The work also helps to use an automated process to alert the risk of adverse health conditions and explore the symptoms related to the detected conditions.



Our proposed model consists of two main components, a rationalising block and the classification block, as shown in Figure \ref{fig:overview}. In a high-level overview, the rational layers select the important time steps and pass to an LSTM layer. The LSTM layer will ignore the trivial time steps and process the data for the attention block. The classifier uses these time points for predictions. After processing by the attention block, the model will give a prediction. The details of these blocks are explained in the following sections.

\subsection*{Positional Encoding}
To use the order of sequence in the analysis, we add positional encoding (PE) before passing the data into the model. We use the sine and cosine positional encoding \cite{vaswani2017attention}. Shown in Equation \ref{eq:pe}, where $pos$ is the position of the time step, $i$ is the position of the sensor, $d$ is the dimension of each time step. 

\begin{equation}
\label{eq:pe}
\begin{split}
    PE(pos, 2i) = sin(pos/10000^{2i/d}) \\
    PE(pos, 2i + 1) = cos(pos/10000^{2i/d})
\end{split}
\end{equation}

\subsection*{Rationalising Prediction}

To add more focus on the time steps in the data that are more relevant to the predictions, the generator produces a binary mask to select or ignore a specific time points. For example: 
$\textbf{x} \in \mathbb{R}^{k \times f}$ contains $k$ time point and $f$ features for each time point, the generator will produce a binary vector $\textbf{z}=\{z_1,z_2,\dots,z_k\}$. The $i^{th}$ variable $z_i \in \{0,1\}$ indicates whether the $i^{th}$ time point in $\textbf{x}$ is selected or not.

Whether the $i^{th}$ time point is selected or not is a conditional probability given the input $x$. We assume that the selection of each time point is independent. The Generator uses a probability distribution over the $\mathbf{z}$, which could be a joint probability of the selections. The joint probability is given by:

\begin{equation}
\label{eq:joint_prob}
p(z|x) = \prod^k_{i=1}p(z_i|x)
\end{equation}

\subsection*{Classifier}
After exploring and selecting the most relevant time points, we train a classifier to provide the predictions. The trained classifier contains attention blocks and residual blocks.

Attention block is an application of self-attention mechanism to detect the important features. The attention mechanism detects important parts of a sequence. It has three key components: the inputs structure, the compatibility function and the distribution function \cite{galassi2019attention}. 

\noindent There are three inputs in the structure; Keys ($\mathbf{K} \in \mathbb{R}^{{n}_{k} \times {d}_{k}}$), Values ($\mathbf{V} \in \mathbb{R}^{{n}_{v} \times {d}_{v}}$) and Query ($\mathbf{Q} \in \mathbb{R}^{{n}_{q}}$), where the $n$ is the dimension of the inputs, the $k, v, q$ are the dimension of the outputs. They could have different or same sources. If $\mathbf{K}$ and $\mathbf{q}$ come from the same source, it is self-attention \cite{vaswani2017attention}. $\mathbf{K}$ and $\mathbf{V}$ represent input sequence which could be either annotated or raw data. $\mathbf{q}$ illustrates the reference sequence for computing attention weights. For combining and comparing the $\mathbf{q}$ and $\mathbf{K}$ values, compatibility function has been used. Distribution function computes the attention weights ($\mathbf{a} \in \mathbb{R}^{{d}_{k}}$) using the output of compatibility function ($\mathbf{c} \in \mathbb{R}^{{d}_{k}}$).

We obtain the attention by Equation \ref{eq:attention}. The $Q, K, V$ are matrices formed by queries, keys and values vectors, respectively. Since we use the self-attention, the $Q, K, V$ are calculated by the inputs with different weight matrices.

\begin{equation}
\label{eq:attention}
\textup{Attention}(Q,K,V) = \textup{softmax}(\frac{QK^T}{\sqrt{d_k}})V
\end{equation}

The architecture of the attention block is the same described in \cite{vaswani2017attention}. We employ a residual connection \cite{he2016deep} followed by a normalisation layer \cite{ba2016layer} inside the attention block. Residual blocks and the output layer process the output of the attention block. 

\subsection*{Objective function}
The training samples in healthcare datasets are often imbalanced due to the low prevalence and sporadic occurrences. In other words, some of the classes contain more samples than others. For example, only 25\% of the data we collected are labelled as positive. More details of the dataset will be clarified in the following section. To deal with the imbalance issue, we use focal loss \cite{lin2017focal} as the objective function of the classifier, shown in Equation \ref{eq:loss_clf}:

\begin{equation}
\label{eq:loss_clf}
\mathit{L_c} = - \alpha(1-p)^\beta \log(p)
\end{equation}

\noindent where $\alpha$ and $\beta$ are hyper-parameters to balance the variant of the focal loss, $p=f(x,z)*y + (1-f(x,z)*(1-y)$. $f(x,z)$ is the probability estimated by the classifier and $y \in \{0,1\}$ is the label of $x$.

In addition to the loss function used in the classifier, the generator produces a short rational selection and calculates the loss. Shown in Equation \ref{eq:loss_gen}, where the $\lambda$ is the parameter to weight the selection:

\begin{equation}
\label{eq:loss_gen}
\mathit{L_g} = \lambda||\textbf{z}||
\end{equation}

We then combine the focal loss and the loss from generator to construct loss function as shown in Equation \ref{eq:loss_combin}:

\begin{equation}
\label{eq:loss_combin}
\mathit{L} = \sum_{(x,y)\in D}\mathbb{E}[\mathit{L_c} + \mathit{L_g}]
\end{equation}

\section{Results}\label{sec:res}

 \begin{figure*}
  \begin{subfigure}{.3\textwidth}
      \centering
     \includegraphics[width=\linewidth]{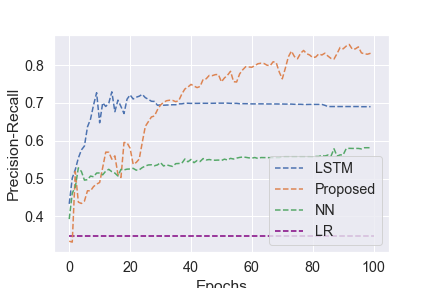}
     \caption{PR}
 \end{subfigure}
 \begin{subfigure}{.3\textwidth}
      \centering
     \includegraphics[width=\linewidth]{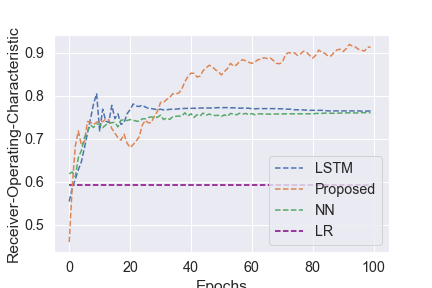}
     \caption{ROC}
 \end{subfigure}
  \begin{subfigure}{.3\textwidth}
      \centering
     \includegraphics[width=\linewidth]{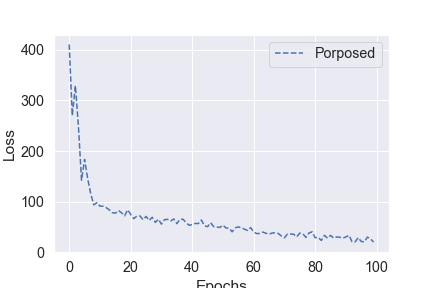}
     \caption{Loss}
     \label{fig:TIHM_loss}
 \end{subfigure}
 \caption{Evaluation of the proposed methods using the in-home sensory dataset. (a) shows the precision; (b) evaluates the Receiver Operating Characteristics (ROC) curve and (c) shows the changes to the loss during the training. In (a) and (b) the results of the proposed model is also compared with a set of baseline models.}
 \label{fig:evaluation_TIHM}
 \end{figure*}

 \begin{figure*}
  \begin{subfigure}{.3\textwidth}
      \centering
     \includegraphics[width=\linewidth]{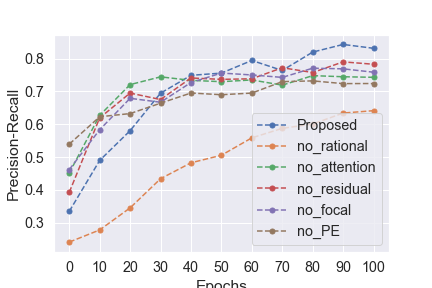}
     \caption{PR}
 \end{subfigure}
 \begin{subfigure}{.3\textwidth}
      \centering
     \includegraphics[width=\linewidth]{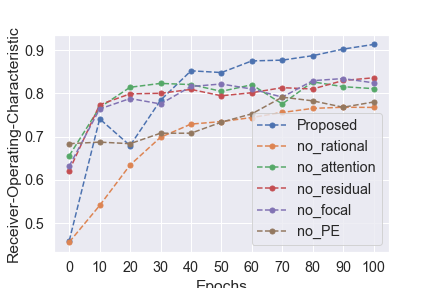}
     \caption{ROC}
 \end{subfigure}
  \begin{subfigure}{.3\textwidth}
      \centering
     \includegraphics[width=\linewidth]{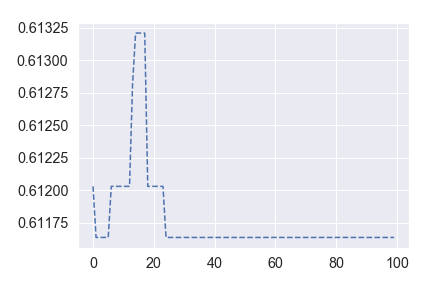}
     \caption{Selection Rate changes}
     \label{fig:TIHM_rate}
 \end{subfigure}
 \caption{An ablation study to evaluate the model; (a) shows the precision; (b) evaluates the Receiver Operating Characteristics (ROC) curve and (c) shows the selection rate changes. In (a) and (b) the results of the evaluation is by eliminating different components from the model. }
 \label{fig:ablation_tihm}
 \end{figure*}

\noindent\textbf{Evaluation Metrics}: To evaluate our proposed method and compare it with the baseline models, we calculated different metrics. One of the primary metrics to assess the model is accuracy which is the measure of how close is the predicted class to the actual class. However, accuracy alone cannot be a good measure to evaluate the performance of a classifier. As a result, we also calculated the Area Under the Curve of Receiver Operating Characteristic (ROC) and Precision-Recall (PR). The precision of class A is the ratio of samples predicted as class A which are correct, and Recall is the ratio of samples as true class A which have been detected. ROC curve is the measure of model capability in differentiating between classes. We do not report the results in terms of specificity and sensitivity. The reason is that in this study, we do not have access to the full electronic healthcare records and hospital admission data of all the participants. So report the specificity and sensitivity only based on the detected and evaluated labels in our dataset, which can only be a sub-set of true and false cases for the cohort, can be misleading in terms of an actual and generalisable clinical finding. Instead, we have opted to evaluate the precision and generalisability of the prediction algorithm based on the existing labelled data and the known cases that we could evaluate and verify the performance of the model.  \\  

\noindent\textbf{Baseline Models}: We compare our model with the Linear Regression (LR) \cite{sperandei2014understanding}, Long-Short Term Memory (LSTM) neural networks \cite{gers1999learning} and a fully connected Neural Network (NN) model \cite{hassoun1995fundamentals}. 

LR is a discriminative model which can avoid the confounding effects by analysing the association of all variables together \cite{sperandei2014understanding}. It is also a commonly used baseline model to evaluate the performance of the proposed models \cite{harutyunyan2019multitask}.

NN has the ability to learn a complex relationship. Unlike LR, NN does not need to assume the variables are linearly separated. It is also applied to a variety of clinical data sets \cite{lasko2013computational, che2015deep}. In the experiment, we used a Neural Network with one hidden layer contains 200 neurons, a softmax output layer contains two neurons, cross entropy loss and adam optimiser.

LSTM is a powerful neural network to analyse the sequential data, including time-wised clinical datasets \cite{choi2016doctor, baytas2017patient}. It can associate the relevant inputs even if they are widely separated. Since our dataset consists of time-series sequences, we take the LSTM as another baseline model. In the experiment, we used a model that contains one residual block, one LSTM layer contains 128 neurons, and a softmax output layer contains two neurons, cross entropy loss and adam optimiser.

In the experiments, we aggregate the readings of each sensor per hour. Hence each data point contains 24-time points and eight features. We set the batch size to $32$, learning rate to $0.0001$, sparsity to $0.001$. We divide the data into a train set and a test set. The numbers of training and testing samples in the datasets are 209 and 103 cases with their associated time-series data, respectively. The data is anonymous, and only the anonymous data without any personally identifiable information is used in this research. \\

\noindent\textbf{Experiments}: The ROC and PR changes during training are shown in the first two graphs in Figure \ref{fig:evaluation_TIHM}. Overall, the proposed model outperforms other baseline methods. The LSTM performs well in dealing with the time-series data. Compared to the other methods, the neural network converges much faster. However, the performance of the model fluctuates around 30 epochs. The convergence and the fluctuation are due to the rational process. The model has to learn how to extract important time steps and pay attention to the features. This process is also reflected in Figure \ref{fig:TIHM_loss}, the loss fluctuates during that period. However, the model adjusts this fluctuation automatically and improves the performance. The overall results are also summarised in Table \ref{tab:results}.

\begin{table}[h!]
    \centering
    \begin{tabular}{|c|c|c|c|c|}
    \hline
         & LR & LSTM & NN & Proposed method\\
     \hline
     AUC - PR   &  0.3472 & 0.6901 & 0.5814 & \textbf{0.8313}\\
     AUC - RC   &  0.5919 & 0.7644 & 0.7601 & \textbf{0.9131}\\
     \hline
    \end{tabular}
    \caption{The evaluation results in comparison with a set of baseline models: Linear Regression (LR), Long-Short Term Memory (LSTM) neural networks and a fully connected Neural Network (NN) model. Since the dataset is imbalance, we calculated the Area Under the Curve (AUC) of Receiver Operating Characteristic (ROC) and Precision-Recall (PR) to evaluate the performance.}
    \label{tab:results}
\end{table}

\section{Discussion}
\textbf{Ablation Study}: We begin the discussion with an ablation study. Our model contains five important components: Rational layers, Attention layers, Residual layers, focal loss and positional encoding. We omit each component one at a time and explore how removing one of the components will impact the performance of the model. The experiments are shown in the first two graphs of Figure \ref{fig:ablation_tihm}.  The orange line represents the model without rational layer. Although the performance of the model without rational layer keeps increasing, it underperforms in others significantly. In other words, the rational layer plays an important role in the model. Removing the positional encoding, attention layer, residual layer, or the focal loss decrease the performance as well. The performance change caused by omitting each of these four components are quite similar. As shown in Figure \ref{fig:ablation_tihm}, the positional encoding helps the model to identify relevant patterns of the data over time and plays an important role in the performance of the model. The rate of selected timesteps changes is shown in Figure \ref{fig:TIHM_rate}.  The rate of selected timesteps changes is shown in Figure \ref{fig:TIHM_rate}. \\

\textbf{Rationalising prediction}: the Rational component helps to increase the accuracy of the model. Generally, the proposed rationalising method shows that the model knows which time steps and features to give the prediction. These patterns and time steps can also be explored to identify and observer relevant data and symptoms to a condition in each patient. Using this component, a personalised set of patterns and symptoms can be explored for each patient. The last graph in Figure \ref{fig:ablation_tihm} shows the selection rate changes during the training phase. The model learns to extract the time steps, and the accuracy increases after the changes become stable. As mentioned in the ablation study, after learning to extract the important time steps, the proposed model outperforms the baseline models without rational mechanisms. In other words, the model extracts a sub-set of the time steps (e.g. part of the time steps are extracted from Figure \ref{fig:visual_agg_data} to Figure \ref{fig:visual_select_data}) to obtain a better prediction. As the learning process continues, the model tries different selections and finds the optimised selection rate. Comparing to other models, the performance of the proposed model does not decrease during the training. The model learns to pay attention to the most relevant segments of the data and consider long-distance dependencies in the time-series data. In summary, the proposed model can not only explain the prediction but also abandon the redundant information in the data automatically. According to our experiments, the proposed model in average selects $61\%$ of the time points in the datasets to estimate the predictions.\\

\begin{figure*}
    \centering
    \includegraphics[width=\linewidth]{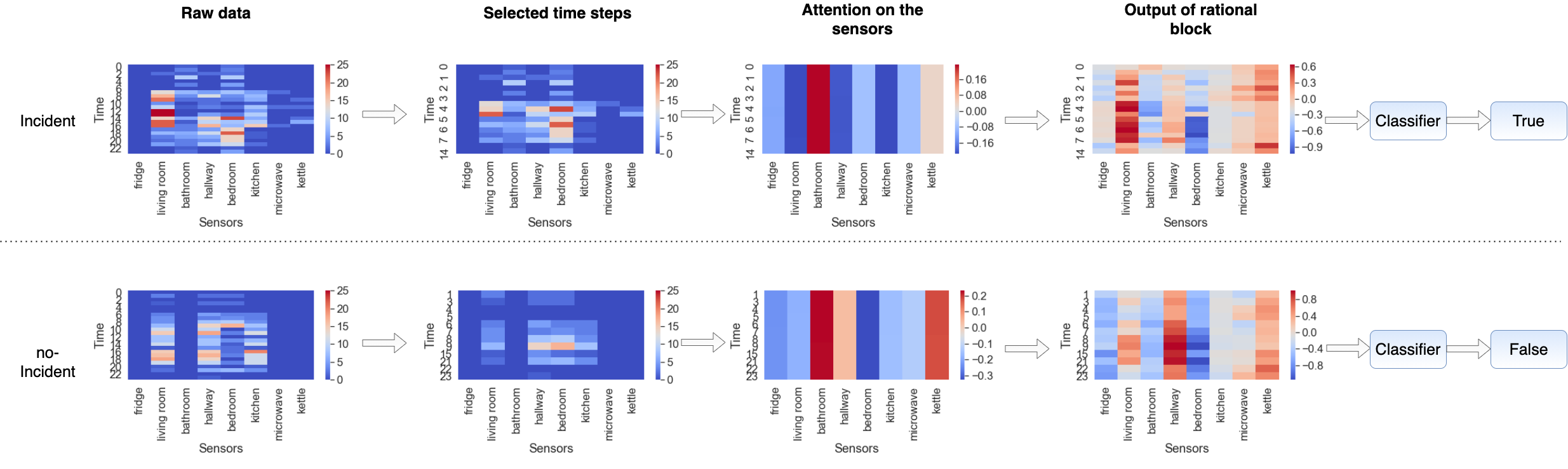}
    \caption{Visualisation of the outputs within the rational block. The top figure visualises a sample which is validated with a True incident. The bottom figure is a sample which is validated with a False incident.}
    \label{fig:rational_pair}
\end{figure*}

\textbf{Pair analysation}: We then analyse the rational block processing on the positive and negative samples. As shown in Figure \ref{fig:rational_pair}, the rational block assigns weights to the positive and negative samples differently. More specifically, the model has learnt to extract different amount and series of time steps based on the inputs. In this case, the model extracts more time steps on the positive case than the negative case. Furthermore, the model pays attention differently based on the input data. In the example above, the model assumes the bathroom is the most important sensors in the positive samples. However, the model takes the bathroom and kettle almost as equally important sensors for predicting the negative case. After the model pays attention to the sensors of selected time steps, the classifier gives the predictions correctly. 

\subsection*{Translating machine learning research into clinical practice}

Improving the quality of life by preventing illness‐related symptoms and negative consequences of dementia has been set out as a major goal to advance dementia care. Agitation and infections have been highlighted as areas for priority development \cite{6buhr2006caregivers}. Our proposed model directly addresses these priorities in dementia care and intervention by enabling early detection of agitation and urinary tract infections in remote healthcare monitoring scenario, providing an opportunity for delivering more personalised, predictive and preventative healthcare. When applied to real-world clinical dataset in the context of the current clinical study our proposed algorithm provided a recall of 91\% and precision of 83\% in detecting early signs of agitation and UTI from physiological and environmental sensor data. A clinical monitoring team verified the predictions by contacting the patient or carer when an agitation or UTI alert was generated. A set of clinical pathways for early interventions has also been developed for the clinical monitoring team to use when responding to the alerts. 

\subsubsection*{Relevance to patient outcomes}

We would like to highlight an important aspect of using this type of analysis to evaluate healthcare and patient outcomes. Focusing only on accuracy as a metric for assessment of the solution within a specific cohort goes only so far \cite{MITNews}. Large studies and further experiments with different cohorts and various in-home deployment settings are required to assess how such algorithms will perform in the noisy and dynamic real-world environments. There are several examples of AI and machine learning algorithms that perform very well in controlled and laboratory settings, but the real-world experience is different \cite{MITNews}. In this study, the sensors and data collection happens in uncontrolled, real-world environment. We have done several cross-validations, comparison and ablation studies to avoid overfitting the model and make sure the results are robust and reproducible. However, further independent trials and validation studies with larger cohorts are required to transform the current work into a product that can be used in real-world clinical and care settings. Another important item is that only focusing on the accuracy of the algorithm will not give a complete picture of the real effectiveness and impact of the solution on patient outcomes.  

Our agitation intervention protocol follows all current guidelines, which agree that individualised and person-centred non-pharmacological therapies are the first-line treatment for agitation in people with dementia \cite{52duff2018dementia, 53ijaopo2017dementia}. In line with the current guidelines, the initial assessment explores possible reasons for patients’ distress and addresses clinical or environmental causes first. The clinical monitoring team asks a set of standardised questions to evaluate the symptoms and to help the carer to identify  potential causes of agitation such as pain, illness, discomfort, hunger, loneliness, boredom or environmental factors (temperature, light, noise level). The recognition and treatment of possible organic causes or triggering factors remains the mainstem of the intervention. In particular detection of delirium and a possible underlying infection is of great importance and the clinical monitoring team  facilitates early diagnosis and treatment by liaising with the study’s clinical team and patient’s GP. Finally, the clinical monitoring team provides psychological support for the caregivers in order to reduce the caregiver distress. In the future, we are planning to use multimodal sensor data to improve the classification of agitation state which will include measuring sound levels along with activity detected by environmental sensors.

Similarly to the agitation protocol, in case of a UTI alert the clinical monitoring team first responds by contacting the patient/carer to evaluate the symptoms. However, the diagnosis of UTI in dementia patients can be problematic, as these patients are less likely to present with a typical clinical history and localised urinary symptoms compared with younger patients \cite{54lutters2008antibiotic}. The team, therefore, arranges a home visit to perform a dipstick urine test. If the urine dipstick test is suggestive of infection (positive nitrates or leukocytes) clinical monitoring team advises the person with dementia/carer to visit the GP the same day to obtain a prescription for antibiotics. Monitoring Team also informs the GP of test results and requesting for antibiotics to be prescribed.

One potential criticism of our UTI intervention algorithm could be the possibility of antibiotic over-prescribing contributing to the spread of antibiotic resistance. However, recent evidence demonstrates that in elderly patients with a diagnosis of UTI in primary care, no antibiotics and delayed antibiotics are associated with a significant increase in bloodstream infection and all-cause mortality compared with immediate treatment \cite{55gharbi2019antibiotic}. Therefore, early prescription of antibiotics for this vulnerable group of older adults is advised in view of their increased susceptibility to sepsis after UTI and despite a growing pressure to reduce inappropriate antibiotic use. 

The impact of our in-home monitoring technologies and the embedded machine learning models on clinical outcomes including hospitalisation, institutionalisation  and mortality rates is part of an ongoing study. Nevertheless, the current work demonstrates the effectiveness of the  proposed algorithm and its translation into real-life clinical interventions. Fig \ref{fig:rational_pair} illustrates individual cases of agitation and UTI correctly identified by the algorithm, with the digital markers demonstrating a behavioural anomaly.

\section{Conclusion}
To avoid unplanned hospital admissions and provide early clues to detect the risk of agitations and infections, we collected the daily activity data and vital signs by in-home sensory devices. The noise and redundant information in the data lead to inaccuracy predictions for the traditional machine learning algorithms. Furthermore, the traditional machine learning models cannot give explanation of the predictions. To address these issues, we proposed a model that can not only outperform the traditional machine learning methods but also provide the explanation of the predictions. The proposed rationalising block, which is based on the rational and attention mechanism, can process healthcare time-series data by filtering the redundant and less informative information. Furthermore, the filtered data can be regarded as the important information to support clinical treatment. We also demonstrate the focal loss can help to improve the performance on the imbalanced clinical dataset and attention-based models can be used effectively in healthcare data analysis. The evaluation shows the effectiveness of the model in a real-world clinical dataset and describes how it is used to support people with dementia. \\

\section*{Acknowledgment}

This research is funded by the UK Medical Research Council (MRC), Alzheimer's Society and Alzheimer's Research UK and supported by the UK Dementia Research Institute.

\bibliographystyle{IEEEtran}
\bibliography{IEEEabrv, sample}

\begin{thebibliography}{10}
\providecommand{\url}[1]{#1}
\csname url@samestyle\endcsname
\providecommand{\newblock}{\relax}
\providecommand{\bibinfo}[2]{#2}
\providecommand{\BIBentrySTDinterwordspacing}{\spaceskip=0pt\relax}
\providecommand{\BIBentryALTinterwordstretchfactor}{4}
\providecommand{\BIBentryALTinterwordspacing}{\spaceskip=\fontdimen2\font plus
\BIBentryALTinterwordstretchfactor\fontdimen3\font minus
  \fontdimen4\font\relax}
\providecommand{\BIBforeignlanguage}[2]{{%
\expandafter\ifx\csname l@#1\endcsname\relax
\typeout{** WARNING: IEEEtran.bst: No hyphenation pattern has been}%
\typeout{** loaded for the language `#1'. Using the pattern for}%
\typeout{** the default language instead.}%
\else
\language=\csname l@#1\endcsname
\fi
#2}}
\providecommand{\BIBdecl}{\relax}
\BIBdecl

\bibitem{world2012dementia}
``World health organization. dementia: a public health priority,''
  \url{https://www.who.int/mental_health/publications/dementia_report_2012/en/},
  2012.

\bibitem{alz}
``Alzheimer’s society: dementia support and research charity,''
  \url{https://www.alzheimers.org.uk/}.

\bibitem{3livingston2020dementia}
G.~Livingston, J.~Huntley, A.~Sommerlad, D.~Ames, C.~Ballard, S.~Banerjee,
  C.~Brayne, A.~Burns, J.~Cohen-Mansfield, C.~Cooper \emph{et~al.}, ``Dementia
  prevention, intervention, and care: 2020 report of the lancet commission,''
  \emph{The Lancet}, vol. 396, no. 10248, pp. 413--446, 2020.

\bibitem{4pickett2018roadmap}
J.~Pickett, C.~Bird, C.~Ballard, S.~Banerjee, C.~Brayne, K.~Cowan, L.~Clare,
  A.~Comas-Herrera, L.~Corner, S.~Daley \emph{et~al.}, ``A roadmap to advance
  dementia research in prevention, diagnosis, intervention, and care by 2025,''
  \emph{International journal of geriatric psychiatry}, vol.~33, no.~7, pp.
  900--906, 2018.

\bibitem{5feast2016behavioural}
A.~Feast, M.~Orrell, G.~Charlesworth, N.~Melunsky, F.~Poland, and
  E.~Moniz-Cook, ``Behavioural and psychological symptoms in dementia and the
  challenges for family carers: systematic review,'' \emph{The British Journal
  of Psychiatry}, vol. 208, no.~5, pp. 429--434, 2016.

\bibitem{6buhr2006caregivers}
G.~T. Buhr, M.~Kuchibhatla, and E.~C. Clipp, ``Caregivers' reasons for nursing
  home placement: clues for improving discussions with families prior to the
  transition,'' \emph{The Gerontologist}, vol.~46, no.~1, pp. 52--61, 2006.

\bibitem{7peach2016risk}
B.~C. Peach, G.~J. Garvan, C.~S. Garvan, and J.~P. Cimiotti, ``Risk factors for
  urosepsis in older adults: a systematic review,'' \emph{Gerontology and
  geriatric medicine}, vol.~2, p. 2333721416638980, 2016.

\bibitem{8tal2005profile}
S.~Tal, V.~Guller, S.~Levi, R.~Bardenstein, D.~Berger, I.~Gurevich, and
  A.~Gurevich, ``Profile and prognosis of febrile elderly patients with
  bacteremic urinary tract infection,'' \emph{Journal of Infection}, vol.~50,
  no.~4, pp. 296--305, 2005.

\bibitem{9fogg2018hospital}
C.~Fogg, P.~Griffiths, P.~Meredith, and J.~Bridges, ``Hospital outcomes of
  older people with cognitive impairment: An integrative review,''
  \emph{International journal of geriatric psychiatry}, vol.~33, no.~9, pp.
  1177--1197, 2018.

\bibitem{Enshaeifar20}
S.~Enshaeifar, P.~Barnaghi, S.~Skillman, D.~Sharp, R.~Nilforooshan, and
  H.~Rostill, ``A digital platform for remote healthcare monitoring,'' in
  \emph{Companion Proceedings of the Web Conference}, 2020.

\bibitem{11majumder2017smart}
S.~Majumder, E.~Aghayi, M.~Noferesti, H.~Memarzadeh-Tehran, T.~Mondal, Z.~Pang,
  and M.~J. Deen, ``Smart homes for elderly healthcare—recent advances and
  research challenges,'' \emph{Sensors}, vol.~17, no.~11, p. 2496, 2017.

\bibitem{12turjamaa2019smart}
R.~Turjamaa, A.~Pehkonen, and M.~Kangasniemi, ``How smart homes are used to
  support older people: An integrative review,'' \emph{International Journal of
  Older People Nursing}, vol.~14, no.~4, p. e12260, 2019.

\bibitem{13peetoom2015literature}
K.~K. Peetoom, M.~A. Lexis, M.~Joore, C.~D. Dirksen, and L.~P. De~Witte,
  ``Literature review on monitoring technologies and their outcomes in
  independently living elderly people,'' \emph{Disability and Rehabilitation:
  Assistive Technology}, vol.~10, no.~4, pp. 271--294, 2015.

\bibitem{miotto2016deep}
R.~Miotto, L.~Li, B.~A. Kidd, and J.~T. Dudley, ``Deep patient: an unsupervised
  representation to predict the future of patients from the electronic health
  records,'' \emph{Scientific reports}, vol.~6, no.~1, pp. 1--10, 2016.

\bibitem{ross2017risk}
C.~S. Ross-Innes, H.~Chettouh, A.~Achilleos, N.~Galeano-Dalmau,
  I.~Debiram-Beecham, S.~MacRae, P.~Fessas, E.~Walker, S.~Varghese, T.~Evan
  \emph{et~al.}, ``Risk stratification of barrett's oesophagus using a
  non-endoscopic sampling method coupled with a biomarker panel: a cohort
  study,'' \emph{The lancet Gastroenterology \& hepatology}, vol.~2, no.~1, pp.
  23--31, 2017.

\bibitem{lipton2015learning}
Z.~C. Lipton, D.~C. Kale, C.~Elkan, and R.~Wetzel, ``Learning to diagnose with
  lstm recurrent neural networks,'' \emph{arXiv preprint arXiv:1511.03677},
  2015.

\bibitem{esteban2016predicting}
C.~Esteban, O.~Staeck, S.~Baier, Y.~Yang, and V.~Tresp, ``Predicting clinical
  events by combining static and dynamic information using recurrent neural
  networks,'' in \emph{2016 IEEE International Conference on Healthcare
  Informatics (ICHI)}.\hskip 1em plus 0.5em minus 0.4em\relax IEEE, 2016, pp.
  93--101.

\bibitem{choi2016doctor}
E.~Choi, M.~T. Bahadori, A.~Schuetz, W.~F. Stewart, and J.~Sun, ``Doctor ai:
  Predicting clinical events via recurrent neural networks,'' in \emph{Machine
  Learning for Healthcare Conference}, 2016, pp. 301--318.

\bibitem{baytas2017patient}
I.~M. Baytas, C.~Xiao, X.~Zhang, F.~Wang, A.~K. Jain, and J.~Zhou, ``Patient
  subtyping via time-aware lstm networks,'' in \emph{Proceedings of the 23rd
  ACM SIGKDD international conference on knowledge discovery and data mining},
  2017, pp. 65--74.

\bibitem{harutyunyan2019multitask}
H.~Harutyunyan, H.~Khachatrian, D.~C. Kale, G.~Ver~Steeg, and A.~Galstyan,
  ``Multitask learning and benchmarking with clinical time series data,''
  \emph{Scientific data}, vol.~6, no.~1, pp. 1--18, 2019.

\bibitem{choi2016retain}
E.~Choi, M.~T. Bahadori, J.~Sun, J.~Kulas, A.~Schuetz, and W.~Stewart,
  ``Retain: An interpretable predictive model for healthcare using reverse time
  attention mechanism,'' in \emph{Advances in Neural Information Processing
  Systems}, 2016, pp. 3504--3512.

\bibitem{ma2017dipole}
F.~Ma, R.~Chitta, J.~Zhou, Q.~You, T.~Sun, and J.~Gao, ``Dipole: Diagnosis
  prediction in healthcare via attention-based bidirectional recurrent neural
  networks,'' in \emph{Proceedings of the 23rd ACM SIGKDD international
  conference on knowledge discovery and data mining}.\hskip 1em plus 0.5em
  minus 0.4em\relax ACM, 2017, pp. 1903--1911.

\bibitem{bai2018interpretable}
T.~Bai, S.~Zhang, B.~L. Egleston, and S.~Vucetic, ``Interpretable
  representation learning for healthcare via capturing disease progression
  through time,'' in \emph{Proceedings of the 24th ACM SIGKDD International
  Conference on Knowledge Discovery \& Data Mining}.\hskip 1em plus 0.5em minus
  0.4em\relax ACM, 2018, pp. 43--51.

\bibitem{ma2019adacare}
L.~Ma, J.~Gao, Y.~Wang, C.~Zhang, J.~Wang, W.~Ruan, W.~Tang, X.~Gao, and X.~Ma,
  ``Adacare: Explainable clinical health status representation learning via
  scale-adaptive feature extraction and recalibration,'' \emph{arXiv preprint
  arXiv:1911.12205}, 2019.

\bibitem{song2018attend}
H.~Song, D.~Rajan, J.~J. Thiagarajan, and A.~Spanias, ``Attend and diagnose:
  Clinical time series analysis using attention models,'' in
  \emph{Thirty-second AAAI conference on artificial intelligence}, 2018.

\bibitem{vaswani2017attention}
A.~Vaswani, N.~Shazeer, N.~Parmar, J.~Uszkoreit, L.~Jones, A.~N. Gomez,
  {\L}.~Kaiser, and I.~Polosukhin, ``Attention is all you need,'' in
  \emph{Advances in neural information processing systems}, 2017, pp.
  5998--6008.

\bibitem{hochreiter2001gradient}
S.~Hochreiter, Y.~Bengio, P.~Frasconi, J.~Schmidhuber \emph{et~al.}, ``Gradient
  flow in recurrent nets: the difficulty of learning long-term dependencies,''
  2001.

\bibitem{johnson2019survey}
J.~M. Johnson and T.~M. Khoshgoftaar, ``Survey on deep learning with class
  imbalance,'' \emph{Journal of Big Data}, vol.~6, no.~1, p.~27, 2019.

\bibitem{jain2019attention}
S.~Jain and B.~C. Wallace, ``Attention is not explanation,'' \emph{arXiv
  preprint arXiv:1902.10186}, 2019.

\bibitem{chawla2002smote}
N.~V. Chawla, K.~W. Bowyer, L.~O. Hall, and W.~P. Kegelmeyer, ``Smote:
  synthetic minority over-sampling technique,'' \emph{Journal of artificial
  intelligence research}, vol.~16, pp. 321--357, 2002.

\bibitem{liu2008exploratory}
X.-Y. Liu, J.~Wu, and Z.-H. Zhou, ``Exploratory undersampling for
  class-imbalance learning,'' \emph{IEEE Transactions on Systems, Man, and
  Cybernetics, Part B (Cybernetics)}, vol.~39, no.~2, pp. 539--550, 2008.

\bibitem{krawczyk2016learning}
B.~Krawczyk, ``Learning from imbalanced data: open challenges and future
  directions,'' \emph{Progress in Artificial Intelligence}, vol.~5, no.~4, pp.
  221--232, 2016.

\bibitem{35lyons2015corrigendum}
B.~E. Lyons, D.~Austin, A.~Seelye, J.~Petersen, J.~Yeargers, T.~Riley,
  N.~Sharma, N.~Mattek, H.~Dodge, K.~Wild \emph{et~al.}, ``Corrigendum:
  Pervasive computing technologies to continuously assess alzheimer's disease
  progression and intervention efficacy,'' \emph{Frontiers in aging
  neuroscience}, vol.~7, p. 232, 2015.

\bibitem{36akl2015autonomous}
A.~Akl, B.~Taati, and A.~Mihailidis, ``Autonomous unobtrusive detection of mild
  cognitive impairment in older adults,'' \emph{IEEE transactions on biomedical
  engineering}, vol.~62, no.~5, pp. 1383--1394, 2015.

\bibitem{37schwickert2013fall}
L.~Schwickert, C.~Becker, U.~Lindemann, C.~Mar{\'e}chal, A.~Bourke, L.~Chiari,
  J.~Helbostad, W.~Zijlstra, K.~Aminian, C.~Todd \emph{et~al.}, ``Fall
  detection with body-worn sensors,'' \emph{Zeitschrift f{\"u}r Gerontologie
  und Geriatrie}, vol.~46, no.~8, pp. 706--719, 2013.

\bibitem{38lazarou2016novel}
I.~Lazarou, A.~Karakostas, T.~G. Stavropoulos, T.~Tsompanidis, G.~Meditskos,
  I.~Kompatsiaris, and M.~Tsolaki, ``A novel and intelligent home monitoring
  system for care support of elders with cognitive impairment,'' \emph{Journal
  of Alzheimer's Disease}, vol.~54, no.~4, pp. 1561--1591, 2016.

\bibitem{39bankole2012validation}
A.~Bankole, M.~Anderson, T.~Smith-Jackson, A.~Knight, K.~Oh, J.~Brantley,
  A.~Barth, and J.~Lach, ``Validation of noninvasive body sensor network
  technology in the detection of agitation in dementia,'' \emph{American
  Journal of Alzheimer's Disease \& Other
  Dementias\textsuperscript{\textregistered}}, vol.~27, no.~5, pp. 346--354,
  2012.

\bibitem{40fleiner2016sensor}
T.~Fleiner, P.~Haussermann, S.~Mellone, and W.~Zijlstra, ``Sensor-based
  assessment of mobility-related behavior in dementia: feasibility and
  relevance in a hospital context,'' \emph{International Psychogeriatrics},
  vol.~28, no.~10, p. 1687, 2016.

\bibitem{41rantz2011using}
M.~J. Rantz, M.~Skubic, R.~J. Koopman, L.~Phillips, G.~L. Alexander, S.~J.
  Miller, and R.~D. Guevara, ``Using sensor networks to detect urinary tract
  infections in older adults,'' in \emph{2011 IEEE 13th International
  Conference on e-Health Networking, Applications and Services}.\hskip 1em plus
  0.5em minus 0.4em\relax IEEE, 2011, pp. 142--149.

\bibitem{42enshaeifar2019machine}
S.~Enshaeifar, A.~Zoha, S.~Skillman, A.~Markides, S.~T. Acton, T.~Elsaleh,
  M.~Kenny, H.~Rostill, R.~Nilforooshan, and P.~Barnaghi, ``Machine learning
  methods for detecting urinary tract infection and analysing daily living
  activities in people with dementia,'' \emph{PloS one}, vol.~14, no.~1, p.
  e0209909, 2019.

\bibitem{wu2020stratified}
C.~Wu and M.~E. Thompson, ``Stratified sampling and cluster sampling,'' in
  \emph{Sampling Theory and Practice}.\hskip 1em plus 0.5em minus 0.4em\relax
  Springer, 2020, pp. 33--56.

\bibitem{aha1996comparative}
D.~W. Aha and R.~L. Bankert, ``A comparative evaluation of sequential feature
  selection algorithms,'' in \emph{Learning from data}.\hskip 1em plus 0.5em
  minus 0.4em\relax Springer, 1996, pp. 199--206.

\bibitem{lin2017focal}
T.-Y. Lin, P.~Goyal, R.~Girshick, K.~He, and P.~Doll{\'a}r, ``Focal loss for
  dense object detection,'' in \emph{Proceedings of the IEEE international
  conference on computer vision}, 2017, pp. 2980--2988.

\bibitem{bahdanau2014neural}
D.~Bahdanau, K.~Cho, and Y.~Bengio, ``Neural machine translation by jointly
  learning to align and translate,'' \emph{arXiv preprint arXiv:1409.0473},
  2014.

\bibitem{usama2020self}
M.~Usama, B.~Ahmad, W.~Xiao, M.~S. Hossain, and G.~Muhammad, ``Self-attention
  based recurrent convolutional neural network for disease prediction using
  healthcare data,'' \emph{Computer methods and programs in biomedicine}, vol.
  190, p. 105191, 2020.

\bibitem{galassi2019attention}
A.~Galassi, M.~Lippi, and P.~Torroni, ``Attention, please! a critical review of
  neural attention models in natural language processing,'' \emph{arXiv
  preprint arXiv:1902.02181}, 2019.

\bibitem{he2016deep}
K.~He, X.~Zhang, S.~Ren, and J.~Sun, ``Deep residual learning for image
  recognition,'' in \emph{Proceedings of the IEEE conference on computer vision
  and pattern recognition}, 2016, pp. 770--778.

\bibitem{ba2016layer}
J.~L. Ba, J.~R. Kiros, and G.~E. Hinton, ``Layer normalization,'' \emph{arXiv
  preprint arXiv:1607.06450}, 2016.

\bibitem{sperandei2014understanding}
S.~Sperandei, ``Understanding logistic regression analysis,'' \emph{Biochemia
  medica: Biochemia medica}, vol.~24, no.~1, pp. 12--18, 2014.

\bibitem{gers1999learning}
F.~A. Gers, J.~Schmidhuber, and F.~Cummins, ``Learning to forget: Continual
  prediction with lstm,'' 1999.

\bibitem{hassoun1995fundamentals}
M.~H. Hassoun \emph{et~al.}, \emph{Fundamentals of artificial neural
  networks}.\hskip 1em plus 0.5em minus 0.4em\relax MIT press, 1995.

\bibitem{lasko2013computational}
T.~A. Lasko, J.~C. Denny, and M.~A. Levy, ``Computational phenotype discovery
  using unsupervised feature learning over noisy, sparse, and irregular
  clinical data,'' \emph{PloS one}, vol.~8, no.~6, 2013.

\bibitem{che2015deep}
Z.~Che, D.~Kale, W.~Li, M.~T. Bahadori, and Y.~Liu, ``Deep computational
  phenotyping,'' in \emph{Proceedings of the 21th ACM SIGKDD International
  Conference on Knowledge Discovery and Data Mining}, 2015, pp. 507--516.

\bibitem{MITNews}
W.~D. Heaven, ``Google’s medical ai was super accurate in a lab. real life
  was a different story,''
  \url{https://www.technologyreview.com/2020/04/27/1000658/\\google-medical-ai-accurate-lab-real-life-clinic-covid-\\diabetes-retina-disease/},
  April 2020.

\bibitem{52duff2018dementia}
C.~Duff \emph{et~al.}, ``Dementia: assessment, management and support for
  people living with dementia and their carers,'' 2018.

\bibitem{53ijaopo2017dementia}
E.~Ijaopo, ``Dementia-related agitation: a review of non-pharmacological
  interventions and analysis of risks and benefits of pharmacotherapy,''
  \emph{Translational psychiatry}, vol.~7, no.~10, pp. e1250--e1250, 2017.

\bibitem{54lutters2008antibiotic}
M.~Lutters and N.~B. Vogt-Ferrier, ``Antibiotic duration for treating
  uncomplicated, symptomatic lower urinary tract infections in elderly women,''
  \emph{Cochrane Database of Systematic Reviews}, no.~3, 2008.

\bibitem{55gharbi2019antibiotic}
M.~Gharbi, J.~H. Drysdale, H.~Lishman, R.~Goudie, M.~Molokhia, A.~P. Johnson,
  A.~H. Holmes, and P.~Aylin, ``Antibiotic management of urinary tract
  infection in elderly patients in primary care and its association with
  bloodstream infections and all cause mortality: population based cohort
  study,'' \emph{bmj}, vol. 364, 2019.

\end{thebibliography}
\end{document}